\documentclass[lettersize,journal]{IEEEtran}
\usepackage{amsmath,amsfonts}
\usepackage{amssymb}
\usepackage{algorithmic}
\usepackage{algorithm}
\usepackage{array}
\usepackage[caption=false,font=normalsize,labelfont=sf,textfont=sf]{subfig}
\usepackage{textcomp}
\usepackage{stfloats}
\usepackage{url}
\usepackage{verbatim}
\usepackage{graphicx}
\usepackage{cite}
\usepackage{booktabs}
\usepackage[pagebackref,breaklinks,colorlinks]{hyperref}
\usepackage{color}
\newcommand{\hxc}[1]{\textcolor[rgb]{0.0,0.0,1.0}{}}

\newcommand{\qcx}[1]{\textcolor[rgb]{1.0,0.0,0.0}{}}

\newcommand{\fqy}[1]{\textcolor{black}{#1}}
\newcommand{\qy}[1]{\textcolor[rgb]{0.2,0.3,0.4}{}}

\newcommand{\tjc}[1]{\textcolor{black}{}}
\newcommand{\tj}[1]{\textcolor{black}{#1}}

\begin{document}

\title{SLS4D: Sparse Latent Space for 4D Novel View Synthesis}

% \author{Qi-Yuan Feng, Hao-Xiang Chen, Qun-Ce Xu, Tai-Jiang Mu, Shi-Min Hu,~\IEEEmembership{Senior Member,~IEEE,}
\author{Qi-Yuan Feng, Hao-Xiang Chen, Qun-Ce Xu, Tai-Jiang Mu
        % <-this % stops a space
\thanks{Qi-Yuan Feng, Hao-Xiang Chen, Qun-Ce Xu, Tai-Jiang Mu are with BNRist, Department of Computer Science and Technology, Tsinghua University, Bejing 100084, China.}% <-this % stops a space
%\thanks{Manuscript received April 19, 2021; revised August 16, 2021.
}



\maketitle
\begin{abstract}
% The ABSTRACT is to be in fully justified italicized text, at the top of the left-hand column, below the author and affiliation information.
% Use the word ``Abstract'' as the title, in 12-point Times, boldface type, centered relative to the column, initially capitalized.
% The abstract is to be in 10-point, single-spaced type.
% Leave two blank lines after the Abstract, then begin the main text.
% Look at previous \confName abstracts to get a feel for style and length.
% Neural radiance fields have played a pivotal role in the field of novel view synthesis and 3D representation. However, the representation in 4D space(e.g., dynamic scene) has been limited by the network representation ability and dynamic matching accuracy. The straightforward solution of using MLP or 3D grid as basic network leads to a huge amount of parameters, and the rendering quality is limited by the size of the grid feature field.
% And directly using of the deformation field to move/match evry 3D points will lead to mismatching of similar points. For the problem of huge parameters but not good enough rendering quality, we prexperimentopose a new implicit grid and a new network interpolation method, achieve better rendering quality in 4D neural radiance field with a feature field of less than 0.2\textperthousand\ in size  and a MLP of similar size. For the problem of inaccurate dynamic filed matching, we propose a cluster matching method, which ensures a beter matching effect, thus further improving the reconstruction effect.

Neural radiance field (NeRF) has achieved great success in novel view synthesis and 3D representation for static scenarios. 
Existing dynamic NeRFs usually exploit a locally dense grid to fit the deformation field; however, they fail to capture the global dynamics and concomitantly yield models of heavy parameters.
We observe that the 4D space is inherently sparse. Firstly, the deformation field is sparse in spatial but dense in temporal due to the continuity of of motion.
Secondly, the radiance field is only valid on the surface of the underlying scene, usually occupying a small fraction of the whole space.
We thus propose to represent the 4D scene using a learnable sparse latent space, a.k.a. SLS4D.
Specifically, SLS4D first uses dense learnable time slot features to depict the temporal space, from which the deformation field is fitted with linear multi-layer perceptions (MLP) to predict the displacement of a 3D position at any time. It then learns the spatial features of a 3D position using another sparse latent space. This is achieved by learning the adaptive weights of each latent code with the attention mechanism.
Extensive experiments demonstrate the effectiveness of our SLS4D:  
it achieves the best 4D novel view synthesis using only about $6\%$ parameters of the most recent work.
\end{abstract}

    
\begin{figure*}
    \centering
    \includegraphics[width=\linewidth]{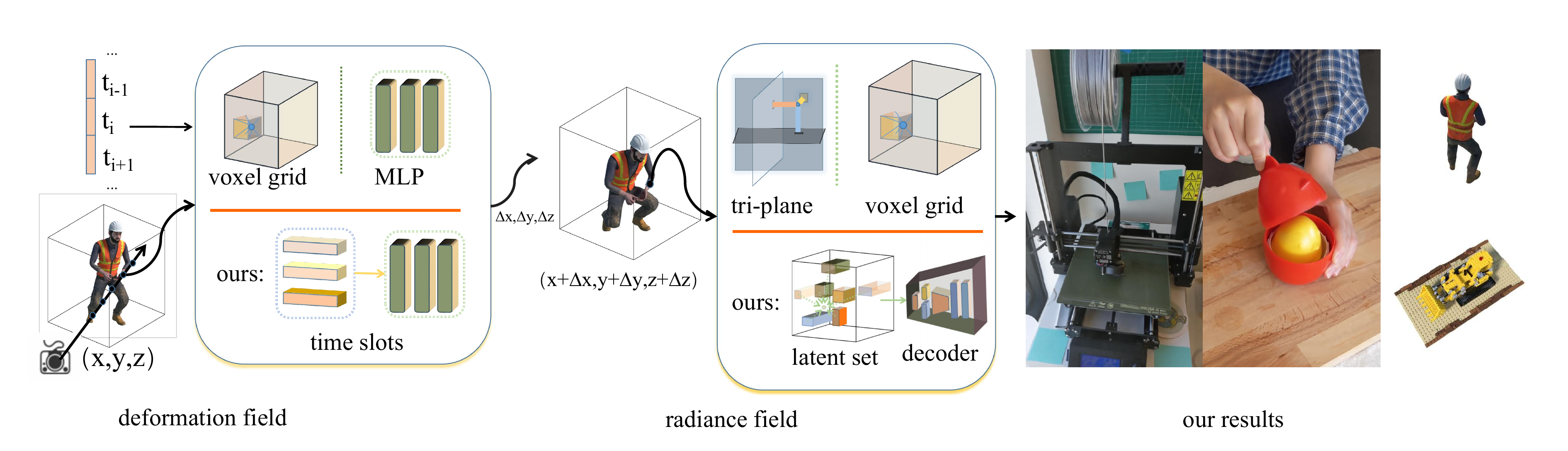}
    \caption{
    Our 4D representation follows a typical dynamic pipeline, which primarily consists of deformation fields and radiance fields. 
    Top: conventionally, the former is represented using voxel grids or MLP decoders, while the latter is decoded using tri-plane or voxel grids. 
    Bottom: We use time slots to more effectively fit the dynamics and characterize the radiance fields in a more compact and informative latent space.
    }
    \label{fig:global-label}
\end{figure*}

\section{Introduction}
\label{sec:intro}

\IEEEPARstart{S}{ynthesizing} a novel view of \tj{a dynamic 3D scene gives the high flexibility to view the scene freely, now serving as a fundamental technique for AR/VR applications and protection of intangible cultural heritage.
Given images captured with known poses for static scenes or objects, neural radiance fields (NeRF)~\cite{mildenhall2021nerf} have achieved great success in novel view synthesis and become a popular implicit 3D representation.
NeRF has also been adapted to represent 4D scenes~\cite{gan2023v4d, cao2023hexplane}, usually accompanied by deformation fields for predicting the displacement of a 3D spatial point at any time~\cite{fang2022fast,shao2023tensor4d}.
}

Many existing dynamic NeRFs for 4D representations are, in fact, striking a balance between novel view synthesis quality and the number of network parameters, either having a large number of model parameters or yielding unsatisfying quality of novel view synthesis.
For example, V4D~\cite{gan2023v4d} achieved the highest rendering quality compared to others, but its substantial number of parameters leads to significantly greater memory consumption compared to others. Methods like Hexplane\cite{cao2023hexplane} and TiNeuVox\cite{fang2022fast} occupy less memory space but render inferior quality.
This leaves a long way to the ultimate goal of achieving the best novel view synthesis quality while reducing the number of model parameters.

\tj{No matter whether employing separate deformation fields for the temporal-spatial decomposition of 4D space or not, most dynamic NeRFs parameterize the feature space of the underlying scene using a uniform and dense grid of fixed resolution, such as the voxel-based feature grid, from which the feature of an arbitrary 3D position is usually linearly interpolated from those of its local surrounding grid points.
Consequently, they are less capable of capturing the global dynamic information of the 4D scene and result in heavy network parameters.
Besides, the view synthesis results are closely related to the grid resolution, i.e., more vivid results would require a finer resolution, but at the cost of a huge increase in network parameters, making the network more difficult to train.
}

\tj{Our key observation is that the 4D space is inherently dense in temporal dimension and sparse in both deformation fields and radiance fields' spatial dimensions.
\fqy{On one hand, the deformation varies along the temporal dimension but remains almost the same in most parts of the spatial dimensions due to the continuity of motion. }
On the other hand, the 3D spatial space is sparse for two reasons.
Firstly, most regions in the 3D space are empty and the representation is only meaningful for locations on the surface of the underlying objects. This indicates that a small amount of 3D volume features is needed when reconstructing the content of 3D radiance fields.
Secondly, both the texture and geometry of the 3D space can be regarded as ``piece-wise constant'', since the appearances of neighboring locations are similar and the surfaces of our daily scenarios are piece-wise planar, such as the desktop, computer monitor, etc. This means that both the appearance and geometry of the 3D spatial space can also be approximated with a finite set of colors and planes, respectively.
}

\tj{Based on the above observation and inspired by the idea of using latent codes to represent a sparse space as explored in sparse coding~\cite{sparsecode06} and dictionary learning~\cite{dictionarylearn, guo2022beyond}, we present \emph{SLS4D}, a sparse latent space for 4D representation.
Our approach adopts the typical deformation field-based dynamic NeRF pipeline as illustrated in Fig.~\ref{fig:global-label}.
The core difference is that we are first to introduce latent code to represent both the deformation fields and the radiance fields.
Both the latent spaces for the deformation fields and the radiance fields are expressed as a finite set of latent codes or bases.
Specifically, we first fit the deformation field with a finite set of globally learnable time slot features. The deformation fields are then built by using linear MLP to predict the displacement of a 3D position at any time.
Similarly, the spatial feature of a 3D position is learned in another latent space, which is also defined as a finite set of globally learnable latent codes.
Considering the complexity of 3D space, we learn the adaptive weights of the spatial position to the latent codes with the attention mechanism~\cite{vaswani2017attention}, which determines the weights of each latent code using the inner product of the query feature of the 3D position and the key value of the latent code.
Notably, the latent code has implicitly learned their positions, making it more adaptive to 4D scenes.
In this way, our SLS4D abandons the direct use of coordinate relationships to determine the final features. Instead, it relies on the relative relationships in the latent feature space, which is more compact and informative.
\fqy{This also makes our approach focus on capturing high-frequency temporal information, aiming to obtain more accurate deformation, thus achieving improved dynamic alignment. At the same time, we focus more on low-frequency and high-frequency components compared to previous works, which also contributes to better dynamic novel-view synthesis quality.}
}

Extensive experiments on two public datasets show that our SLS4D has achieved state-of-the-art dynamic novel view synthesis results compared to previous works.
Specifically, we achieve higher quality rendering results while utilizing approximately only $6\%$ of the parameters in previous SOTA work, i.e., V4D~\cite{gan2023v4d}.

To sum up, this paper makes the following contributions:

\begin{itemize}
    \item The first sparse latent space for 4D representation, which significantly reduces the number of network parameters required for dynamic NeRF.
    \item A spatial latent feature space that learns the adaptive weights with attention, allowing the model to integrate more global priors and achieving superior rendering quality.
    \item A time slot that encodes temporal information more effectively, improving the precision of dynamic scene representation.

\end{itemize}
\section{Related Work}
\label{sec:formatting}

\subsection{Neural Rendering for Dynamic Scenes}

Neural networks for 3D spatial view synthesis\cite{mildenhall2021nerf, chen2022tensorf, muller2022instant, zhang2020nerf++, chen2023factor, wang2021nerf, zhang2022nerfusion} and 3D representations\cite{peng2020convolutional, kerbl20233d, yu2021plenoctrees, neural3dgeometrypriors} offers high-resolution and high-precision reconstruction of 3D scenes, compared with traditional rendering approaches. This advancement enables the generation of novel perspectives on 3D scenes with exceptional detail and precision. 

In the realm of 3D representations, different individuals may have diverse preferences, with some prioritizing the quality of representation, as highlighted in the works\cite{barron2021mip, barron2022mip}, while others may lean towards minimizing the network parameters, as exemplified in the efficient NeRF work~\cite{hu2022efficientnerf, muller2022instant}. There are also those who place a premium on training speed, as seen in the research ~\cite{muller2022instant, chen2022tensorf}.

\tj{
Existing dynamic 3D representations fall into explicit representations and implicit representations.
Explicit representations~\cite{batina1991unsteady,huo2010layered,wu20234d} are friendly to traditional processing/rendering pipelines and can be very efficient, but exhibit a direct correlation between parameter numbers and rendering quality, making optimization less meaningful. 
}

For implicit dynamic 3D representations, some works tend to give precedence to speed like~\cite{fang2022fast,cao2023hexplane, liu2022devrf}, but the pursuit of rendering quality remains a central focus for nearly all works~\cite{pumarola2021d, park2021nerfies,
fang2022fast, song2023nerfplayer, 
shao2023tensor4d, li2022neural, cao2023hexplane, fridovich2023k, gan2023v4d,
gao2021dynamic, li2021neural}.

% \fqy{the following paragraph should be transformed to related work part.}
Many of the foundational approaches for dynamic 3D implicit representations are based on NeRF\cite{mildenhall2021nerf}. We can categorize methods into three types: those that utilize deformation fields with unfixed input camera pose\cite{pumarola2021d, park2021nerfies,fang2022fast, song2023nerfplayer, shao2023tensor4d, jang2022d}, those do not use deformation fields with unfixed input camera pose\cite{li2022neural, cao2023hexplane, fridovich2023k, gan2023v4d, park2021hypernerf}
and based on flows with fixed camera pose(\cite{gao2021dynamic,li2021neural}). 

D-NeRF\cite{pumarola2021d} first introduced the concept of deformation fields, modeling 3D dynamic scenes as a combination of deformation fields and radiance fields. This approach significantly enhances the precision of dynamic scene point matching, consequently improving the rendering quality of dynamic 3D objects. The concept of deformation has also served as inspiration for many other methods\cite{pumarola2021d, park2021nerfies,fang2022fast,song2023nerfplayer,shao2023tensor4d}. D-NeRF\cite{pumarola2021d},TiNeuVox\cite{fang2022fast}, Nerfies\cite{park2021nerfies}, NerfPlayer\cite{song2023nerfplayer} construct the deformation network directly using MLP. Tensor4D~\cite{shao2023tensor4d} established nine planes as the deformation fields during the training phase.

In the meantime, designs like HyperNeRF~\cite{park2021hypernerf} and Nerfies~\cite{park2021nerfies} have incorporated the utilization of deformation latent codes~\cite{bojanowski2017optimizing} \fqy{of each input image} to capture temporal effects at distinct time points, yielding encouraging outcomes. Furthermore, they have introduced latent appearance codes in the 3D space, which, in turn, has enhanced the overall effect. 
This approach may lack strong generalization and is challenging to optimize based on the continuity of the timeline.

TiNeuVox\cite{fang2022fast} leverages a multi-layer spatial grid approach to achieve swift convergence in estimating the general contours of the target shape. By employing voxel grids, TiNeuVox effectively narrows down the search space and facilitates the efficient reconstruction of complex geometries. Since the introduction of TensoRF's 3D decomposition approach, numerous methods for representing dynamic models through 3D decomposition have emerged\cite{cao2023hexplane,shao2023tensor4d,fridovich2023k}.

\subsection{3D Representation by Discretization}

Voxels, as a 3D intuitive representation, are widely used for representing 3D scenes\cite{brock2016generative, choy20163d, dai2017shape}. Its advantage lies in its ability to rapidly perform interpolation. However, it often lacks global information, focusing primarily on local content, which can lead to artifacts. The TVLoss\cite{niemeyer2022regnerf} ensures smoothness and reduces artifacts. Nevertheless, the rendering quality is closely related to grid size, and this approach often requires a large number of parameters.
Because of the substantial number of parameters and the prevalence of extensive empty regions, the utilization voxel-based representations is rather insufficient. Hence, several approaches attempted to address this issue by utilizing Octree structures\cite{hane2017hierarchical,riegler2017octnetfusion, wang2017cnn} or employing hash-based methods\cite{dai2020sg} to speed up the process.
Some researchers employ other strategies to represent 3D space\cite{mescheder2019occupancy, park2019deepsdf}. Moreover, recognizing the limitation of regular grid-based representations, irregular grids\cite{yan2022shapeformer, zhang20223dilg} was introduced as a new form of representation. The drawback of this method is that it is still too slow to find the neighboring irregular points. 
\tj{Instead of interpolating feature from the regular grid, we learn the feature representaion in a latent space, which is more compact and informative.}

\subsection{Adaptive Feature Processing by Attention}

Attention mechanisms~\cite{vaswani2017attention} have been widely used in natural language processing and computer vision~\cite{vit} to learn effective feature representation. For spatial attention, the adaptive feature weights generated by attention mechanisms can enhance spatial transformation capabilitis\cite{jaderberg2015spatial}. 
Point-wise attention can capture long-range contextual information\cite{zhao2018psanet, pct21}, such as using triplet attention enables interactions between different feature spaces\cite{misra2021rotate} and using adaptive point cloud feature interpolation to create iso-surface\cite{zhang20233dshape2vecset}.
\tj{In this paper, we first introduce the sparse latent space for 4D novel-view synthesis, which can adaptively learn the for better rendering quality, while significantly reducing the network parameters.}
\section{Preliminary}

\subsection{Neural Radiance Fields for 3D Representation}
Given a set of images with known camera poses for a static scene,
the neural radiance fields, as proposed by Mildenhall et al.~\cite{mildenhall2021nerf}, implicitly reconstruct the spatial density $\sigma$ and colors $c$ observed at position %leveraging the coordinate 
$(x, y, z)$ along the view direction $\boldsymbol{d}$, using a neural network $\Phi_c$ as follows:

\begin{equation}
    (c, \sigma) = \Phi_c(x,y,z,\boldsymbol{d})
\end{equation}

Given the camera position $\boldsymbol{o}$ and and view direction $\boldsymbol{d}$, points can be sampled along the viewing ray $\boldsymbol{r}$: $(x_i, y_i, z_i) = \boldsymbol{o} + \tau_i\boldsymbol{d}$, where $\tau_i$ is the distance from $p_i$  to the camera along the view direction. Then the color of the ray can be obtained by accumulating colors $c_i$ of N sampling points along the ray~\cite{kajiya1984ray,max1995optical,mildenhall2019local}:
\begin{subequations}
\begin{align}\label{eq:vr}
    \hat{C}(\boldsymbol{r}) &= \sum_{i=1}^{N} T_i(1-\exp(-\sigma_i\delta_i))c_i, \\
    T_i &= \exp(-\sum_{j=1}^{i-1}\sigma_j\delta_j), \\
    \delta_i &= \tau_{i+1} - \tau_{i}
\end{align}
\end{subequations}
where $\delta_i$ denotes the distance between two consecutive sampling points.
The novel view is then synthesized by computing the colors for all the rays in the view.

\begin{figure}[t]
    \centering
    \includegraphics[width=\linewidth]{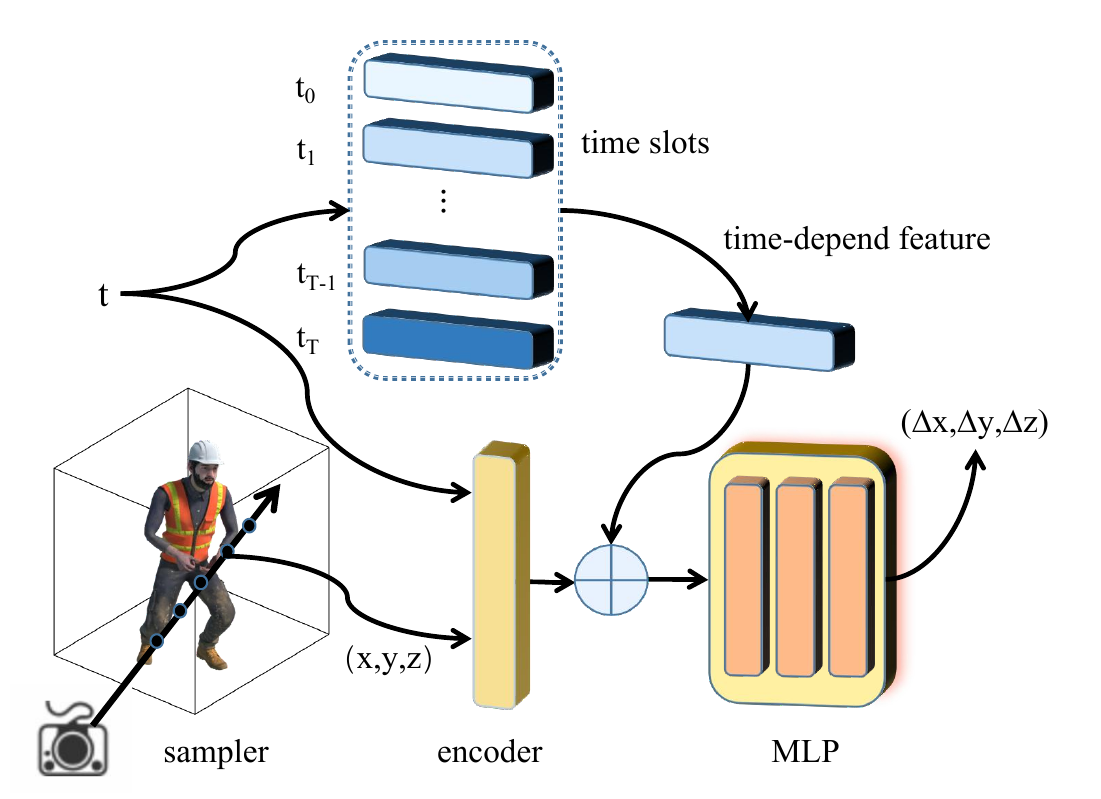}
    \caption{
    \tj{Deformation fields using the latent time slots. The time-depend feature is interpolated from the learnable latent time slots and then decoded into the final deformation $(\Delta x, \Delta y, \Delta z)$ for the input $(x,y,z,t)$.}
    }
    \label{fig:deform-label}
\end{figure}

\subsection{Dynamic 3D Neural Radiance Fields}
Dynamic neural implicit fields often involve concatenating a 3D deformation fields $\Phi_d$ before regular implicit radiance fields $\Phi_c$ (see Fig.~\ref{fig:global-label}). For modeling implicit fields, most works employ MLP or directly interpolate features within the 3D feature space to acquire information. The entire dynamic neural network can be modeled as:
\begin{subequations}
\begin{align}
    (\Delta x,\Delta y,\Delta z) &= \Phi_d(x,y,z,t), \\
    (c,\sigma) &= \Phi_c(x+\Delta x,y+\Delta y,z+\Delta z, \boldsymbol{d})
\end{align}
\end{subequations}
where $(\Delta x,\Delta y,\Delta z)$ is the displacement of point $(x,y, z)$ at time $t$.
Once the color $c$ and density $\sigma$ are known, the final color of a ray can be computed via volume rendering as in Equation~\ref{eq:vr}.
\section{Method}
\label{sec:method}

Our overall pipeline, illustrated in Fig.~\ref{fig:global-label}, first predicts a 3D deformation fields and then reconstructs the radiance fields, following D-NeRF\cite{pumarola2021d}. 
Our core modules include time slot-based deformation fields (Section~\ref{sec:timeslot}) to better capture the high-frequency dynamic information in the temporal dimension, and a latent space, which better leverages the sparse nature of the 3D space to learn more effective radiance fields (Section~\ref{sec:latentspace}). Finally, we detail how to optimize the whole network (Section~\ref{sec:opt}).

\subsection{Latent Time Slots for Deformation Field}
\label{sec:timeslot}

\begin{figure*}
    \centering
    \includegraphics[width=\linewidth]{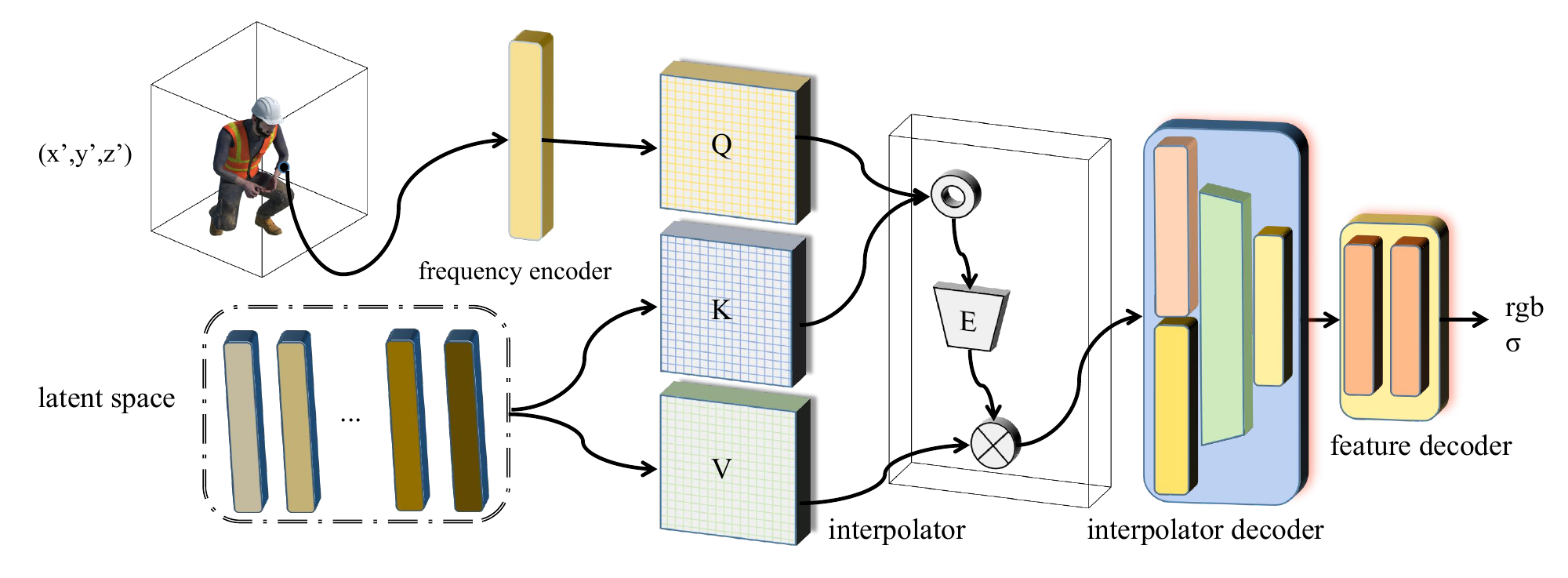}
    \caption{
    Latent space for radiance fields.
    The feature of a 3D position is interpolated in the latent space using the attention mechanism.
    Subsequently, a decoding module and a small MLP are applied to get the final output radiance, i.e., color and density. 
    }
    \label{fig:trans-label}
\end{figure*}

\tj{Instead of directly learning the deformation fields on the spatial grid,}
deformation latent code~\cite{bojanowski2017optimizing} was introduced. 
This method \tj{can produce finer motion but it is time-consuming.} 
In contrast, the Tensor4D~\cite{shao2023tensor4d} directly employed nine planes to interpolate features. While this approach was fast, it consumed a substantial amount of space for interpolation.
\tj{Some works~\cite{park2021nerfies, park2021hypernerf} introduced per-image latent deformation code, but they were less generalizable.}

\tj{Noting that the deformation fields vary along the temporal dimension but remain almost the same in most parts of the spatial dimensions due to the continuity of motion, we employ a slightly dense set of learnable latent time slots $\{\boldsymbol{h}_i\}_0^T$ to generate finer motion of high-frequency.}
\tj{As shown in Fig.~\ref{fig:deform-label},} for the timestamp $t$, we obtain the corresponding time-depend high-frequency feature $\boldsymbol{h}_t$ through linear interpolation. 
This time embedding not only reduces the required space like Tensor4D~\cite{shao2023tensor4d} but also preserves the high-frequency information at each timestamp $t$. 
\tj{Similar to D-NeRF~\cite{pumarola2021d} or TiNeuVox~\cite{fang2022fast}, the input $(x,y,z,t)$ is also encoded into a higher dimension and then concatenated with the time-depend feature $\boldsymbol{h}_t$, which is decoded as the final deformation with an MLP.
We use $\tanh$ as the activation function since it can compress deformations to the range of $[-1,1]$.
}

To ensure temporal continuity over time, we introduce 1D time total variation loss(TV loss) $L_t$ for the latent time slots:
\begin{equation}
    L_{t} = \sum_{i=0}^{T-1}||\boldsymbol{h}_{i+1}-\boldsymbol{h}_{i}||.
\end{equation}

\subsection{Sparse Latent Feature Space for NeRF}
\label{sec:latentspace}

\tj{Having the deformation of a 3D position at any time, now we learn its color and density with radiance fields.
}

\fqy{For 3D radiance fields,} the contemporary discretization-based methods parameterize the neural fields by discretizing the space into a dense regular grid $\mathcal{V}$, where each grid point feature consists of a latent code $\boldsymbol{f}_v$ and its position $v \in \mathcal{V}$. \fqy{This approach neglects the sparsity of the 3D radiance fields, which we will delve deeper into in Section~\ref{sec:disccus}.}
When querying a feature on any position $p$ in the fields, these methods find the neighbor points $\mathcal{N} = \{p_i|p_i \in \mathcal{V}\}$ of $p$ and calculate the weight $w_i$ of the feature $\boldsymbol{f}_{p_i}$, then feature $\boldsymbol{f}_p$ of position $p$ is calculate by the following equation:
\begin{equation}\label{eq:plain-interpolation}
   \boldsymbol{f}_p = \sum_{p_i \in \mathcal{N}} w_i*\boldsymbol{f}_{p_i}.
\end{equation}

In trilinear or bilinear interpolation, $w_i$ is solely related to the distances between the target point and its nearest surrounding four or eight points. These methods are generalizable to different situations but the fitting ability is highly related to the resolution of dense grids, bringing a trade-off between fitting ability and the number of network parameters.
The number of network parameters of dense grids is extremely high when we discretize the 4D space.
For 4D space grids with spatial grid size $N$, temporal grid size $T$, and feature size $F$, the space complexity is $O(N^3TF)$. Similarly, for other representations involving 3D space along with deformation fields, the space complexity is also on the order of $O(N^3F)$, and even decompose-based models, like HexPlane~\cite{cao2023hexplane}, still need $O(3*((N^2 + NT)F))$.

\tj{To better make use of the sparsity of 3D radiance fields and yield a more effective radiance field, as shown in Fig.~\ref{fig:trans-label}, we propose to learn the radiance fields in a more compact and informative space, i.e., the learnable sparse latent space, which also reduces the number of network parameters, making it easier to be trained.
Specifically, we pre-define a set of learnable and global features $\mathbb{V}_s = \{\boldsymbol{f}_1,\boldsymbol{f}_2,...,\boldsymbol{f}_B\} \in R^{F\times B}$, where $B$ is the number of feature vectors and $B << N^2$.
Since this latent feature set has implicitly encoded the position of the feature, for a given 3D position, its feature should be ``interpolated'' according to the distance in latent space instead of the distance in the original 3D space.
To compute how an input 3D position is affiliated to each latent code, we apply the popular attention mechanism~\cite{vaswani2017attention,misra2021rotate,jaderberg2015spatial,hu2018squeeze,zhang20233dshape2vecset} to achieve this:}
\begin{equation}\label{eq:trans-interpolation}
    \boldsymbol{f}_p = \frac{E(\text{Q}(p)^\intercal \text{K}(\mathcal{V}_s))\text{V}(\mathcal{V}_s)}{\sqrt{d}},
\end{equation}
where $\text{Q}(), \text{K}(), \text{V}()$ are responsible for producing the query, key and value vectors, respectively; $d$ is the dimension of the feature and $E(\cdot)$ means the softmax function.  $\text{K}()$ and $\text{V}()$ are linear projections as in the original attention~\cite{vaswani2017attention}.

In the attention module, Q represents positional information related to the target that needs to be queried. We can input the coordinates $(x,y,z)$ or coordinates and time $(x,y,z,t)$ for querying into Q. Multi-head attentions are also applied to enhance the presentation capability of the network.

\fqy{In the case of bilinear or trilinear interpolation in space, we cannot use a frequency encoder to simultaneously focus on both low-frequency and high-frequency information. We use the frequency encoder to produce the query vector, which concentrates both on low-frequency and high-frequency components. The inner product serves as the similarity measure in both the feature space and the frequency domain space.} 

Moreover, when considering the interpolation module, we pass frequency-encoded input through a small projection matrix to perform a transformation from the coordinate frequency space to the dot-product space.

In summary, the preprocessing for $q(x)$ is as follows:
\begin{equation}
\begin{aligned}
    \text{q}(p) = \phi_q((&\sin(2^0p),\cos(2^0p), \\
                          &..., \\
                          &\sin(2^{L-1}p),\cos(2^{L-1}p)))
\end{aligned}
\end{equation}
where $L$ is the maximum frequency and $\phi_q$ is a linear projection.

\tj{The interpolated feature from the latent space is finally fed to a decoding module and a small MLP to produce the output radiance, i.e., color and density.}

In the final part of the network, we first pass the features interpolated from the latent space through a small interpolator decoder. 
We add Gated Linear Units(GEGLU)\cite{shazeer2020glu} to the second layer of the interpolator decoder, reducing the parameter number and improving the learning capability of the interpolate module: 

\begin{align}
    \text{GeGLU}(x, W, V, b, c) &= \text{GELU}(xW+b) \otimes (xV + c) \\
    \text{GELU}(x) &= x \cdot \frac{1}{2}[1+\text{erf}(\frac{x}{\sqrt{2}})]
\end{align}

In this context, we simplify GeGLU by letting $W$, $V$, $b$ and $c$ be optimizable, and erf represents the error function. According to Noam Shazeer\cite{shazeer2020glu}, this approach can improve convergence speed. Finally, we use another small MLP to decode features and output results. 

Similar to other feature-based networks, we separately apply the network described above to the density and color channels and feed the viewpoint information into the MLP.

\subsection{Optimization}
\label{sec:opt}

Unlike in the case of 2D transformers or non-rendering 3D models, we require a more intricate learning rate schedule function $F_{sc}$ to enhance the learning efficiency of the neural network.
After the warm-up phase, we found that simultaneously applying exponential decay and cyclic cosine decay can yield quite favorable results.
The factor of the learning rate is computed by the following equation:
\begin{equation}
    \left\{
    \begin{aligned}
        &\frac{s}{N_{w}} & s \le N_{w}  \\
        &\exp(\frac{s-N_{w}}{N_{m}}) * (1 + \cos(\pi* \frac{s-N_{w}}{N_{m}})) & s > N_{w}
    \end{aligned}
    \right.
\end{equation}
where $s$ is the current training step, $N_{w}$ is warm-up step, and $N_{m}$ is the maximum training step.

We compute L2 loss for each ray and the overall optimization objective $L$ is:
\begin{equation}
    L = w_c * \frac{1}{|R|}\sum_{r \in R}||C(r)-\hat{C}(r)||^2_2 + w_t * L_t
\end{equation}
where $R$ represents the set of rays, $C(r)$ denotes the ground-truth color for ray $r$,  $w_c$ and $w_t$ are balancing weights.

\section{Experiments}

\begin{figure*}
    \centering
    \includegraphics[width=\linewidth]{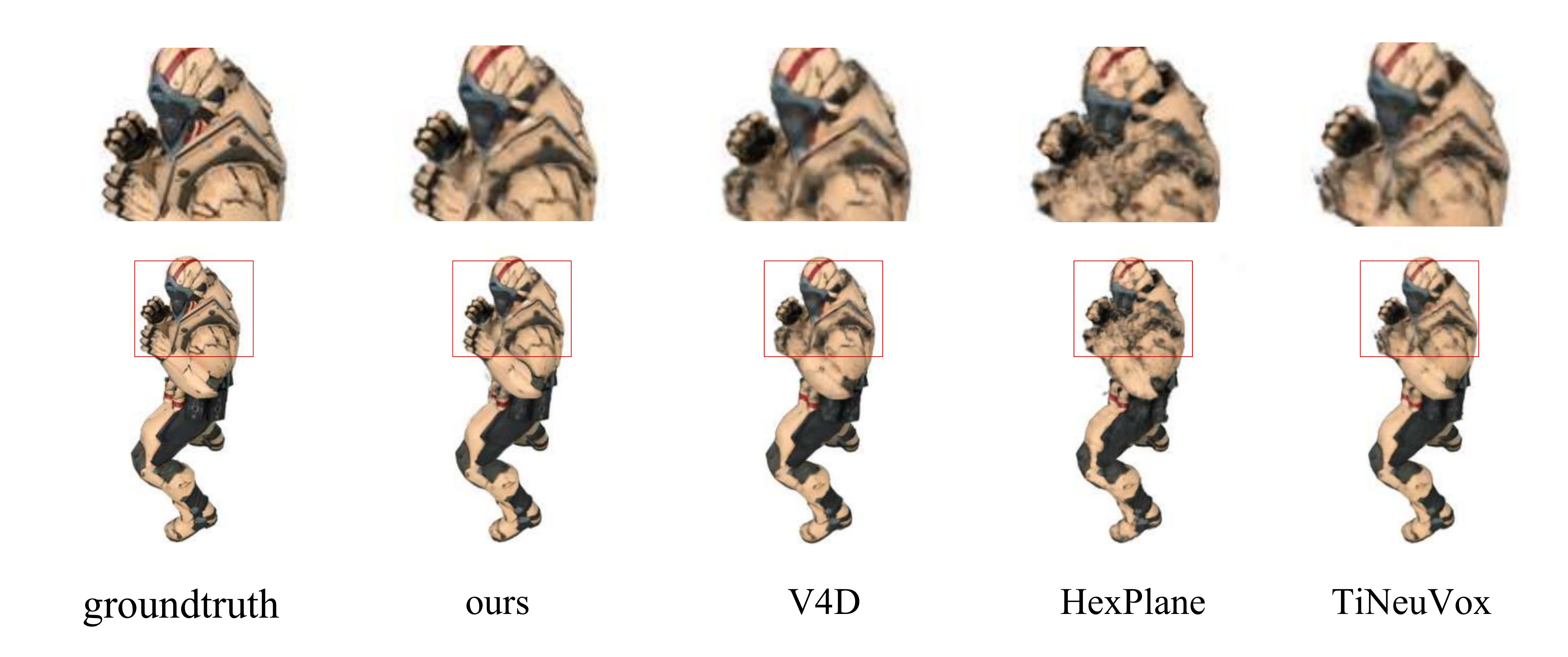}
    \caption{Vsual comparison on the sythetic D-NeRF dataset~\cite{pumarola2021d}. As indicated in the red boxes, 
    Our method achieves the highest clarity and accuracy for both the dynamic parts (i.e., hands) and the static parts (i.e., head and body).
    }
    \label{fig:compare-syn}
\end{figure*}

We conduct experiments on both synthetic (i.e., D-NeRF~\cite{pumarola2021d}) and real-world (i.e., HyperNeRF~\cite{park2021hypernerf}) datasets to evaluate the effectiveness and superiority of our method.

\subsection{\tj{Datasets and Metrics}}

\tj{\textbf{D-NeRF~\cite{pumarola2021d}} provides 8 synthetic dynamic scenes. Each scene has $50 - 150$ training images captured from different camera poses and time steps and 20 images to test.
We follow the official split of train/validation/test to train our model. 
}

\tj{\textbf{HyperNeRF~\cite{park2021hypernerf}} contains 4 dynamic scene captured from the real-world. Each scene has $163-512$ images. We also follow the official split of train/validation/test to train our model.
}

\tj{\textbf{Metrics.} 
To evaluate the dynamic novel-view synthesis quality, we adopt the following recognized metrics:
(1) Peak signal-to-noise ratio (PSNR), (2) Structural similarity index measure(SSIM)~\cite{wang2004image}, (3) Perceptual quality measure (LPIPS)~\cite{zhang2018unreasonable} and (4) Multi-scale structural similarity index measure(MS-SSIM)~\cite{wang2004image}.
A larger value in PSNR, SSIM and MS-SSIM indicates a better dynamic novel-view synthesis quality, while a smaller LPIPS means better results.
}                                                                                                                                                                                   

\subsection{\tj{Implementation Details}}

\tj{To train our model on the D-NeRF dataset}, we set the number of global latent features ($B$) to $256$ for the radiance network, and the dimension of the feature vectors $F$ is $64$. 
Our transformer for the attention consists of $8$ heads with a context dimension of $32$.
\tj{Our interpolator uses a $4$-layer MLP to map the interpolated feature to a dimension of $512$, which is finally decoded to the color and density with a 1-layer and 3-layer MLP, respectively.}
Our deformation fields are similar to other models, comprising a 7-layer MLP along with the initial time-dependent deformation latent code. \fqy{The time slots size $T$ is $256$. During optimization, the weights of $w_c$ and  $w_t$ are set to $1$ and  $0.0001$, respectively.} 
The learning rate is set to $0.0001$ for the feature space, interpolation transformer, and the first three layers of MLP, and $0.001$ for the subsequent layers of MLP 
We trained the network for $N_m=30\text{k}$ iterations to achieve the current results\tj{, with the warm-up step $N_m$ set to $2000$}. 
We employ a strategy similar to the ``empty grid'' approach in Hexplane~\cite{cao2023hexplane} to focus training on specific parts of the target object. 

\tjc{\textbf{All} the configuration of hyper-parameters should be given, such as $T$, $L$ and $w_c$, $w_t$ in the total loss.}

\tj{To train our model on the HyperNeRF dataset,}
we used the same parameters as in the synthetic D-NeRF~\cite{pumarola2021d} dataset, with the only difference being that we double the number of latent features by setting $B=512, T=512$, \fqy{because the total time of scenes on this dataset has increased several times compared to that of D-NeRF dataset.}

All our experiments were conducted on a server of Nvidia 4090.

\subsection{Comparison to Other Methods}

\begin{table}
  \centering
  \caption{Results on the dynamic synthesis dataset of D-NeRF~\cite{pumarola2021d}. Our network occupies an extremely small feature vector space(FS), thus having a smaller parameter number(PN). The best results are in boldface.}
  \label{tab:d-nerf works compare}
  \begin{tabular}{@{}lcccccccccccccccc@{}}
    \toprule
    Method & PSNR$\uparrow$ & SSIM$\uparrow$ & LPIPS$\downarrow$ & FS$\downarrow$ & PN$\downarrow$\\
    \midrule
    T-NeRF\cite{pumarola2021d} &29.51&0.951&0.078&-&$\sim$\textbf{0.1M}\\
    D-NeRF\cite{pumarola2021d} &30.43&0.952&0.066&-&$\sim$0.25M\\
    TiNeuVox-S\cite{fang2022fast} &30.75&0.955&0.066&$\sim$2M&$\sim$2M\\
    TiNeuVox-B\cite{fang2022fast} &32.67&0.973&0.043&$\sim$12M&$\sim$12M\\
    HexPlane\cite{cao2023hexplane} &31.04&0.967& 0.039&$\sim$14M&$\sim$15M\\
    K-Plane\cite{fridovich2023k} & 29.70 &0.961& 0.049&$\sim0.6$M&$\sim$37M\\
    V4D\cite{gan2023v4d} &33.47&0.979&0.028&$\sim$100M&\textgreater100M\\
    SLS4D(ours) &\textbf{34.84}&\textbf{0.981}&\textbf{0.025}&$\sim$\textbf{0.01M}&$\sim$6M \\
    \bottomrule
  \end{tabular}
\end{table}

We compare our method with various models that were previously tested on the D-NeRF~\cite{pumarola2021d} dataset and the results are listed in Table~\ref{tab:d-nerf works compare}. 
As we can see, our method achieved SOTA performance for all three nove-view quality metrics. 
\tj{We also reported the size of feature vector space (FS) and the number of network parameters (PN).}
As we can see, our approach achieves a significant reduction in the number of network parameters compared to other models.
\tj{Notably, our method took $3$ hours to achieve results similar to V4D~\cite{gan2023v4d}, which needs about $7$ hours on Nvidia 4090, and a total of $6$ hours was took to reach the results in Table~\ref{tab:d-nerf works compare}  for our method.}

T-NeRF~\cite{pumarola2021d} directly utilizes a four-dimensional coordinate with an MLP to fit dynamic 3D scenes. D-NeRF~\cite{pumarola2021d} builds upon T-NeRF by introducing deformation fields, separating the time and spatial dimensions.
Other models improve upon D-NeRF because they use more advanced semi-explicit feature space. 
TiNeuVox~\cite{fang2022fast} utilizes hierarchical voxel grids, offering higher speed and the hierarchical grids structure can better capture the frequency information. 
K-Plane~\cite{fridovich2023k} and HexPlane~\cite{cao2023hexplane} employ dimension decomposition to break down the 4D grids into multiple lower-dimensional grid features. V4D~\cite{gan2023v4d} uses 3D grids for interpolation. These methods focus on grid compression and interpolation. As a result, the dynamic novel-view synthesis quality is closely related to the final grid resolution, and they can occupy a considerable amount of network parameters.
\tj{Benefiting from our sparse latent feature space, our method can better represent the 4D space with much fewer network parameters compared to most recent work, i.e., V4D~\cite{gan2023v4d}.
Some visual comparison results are illustrated in Fig.~\ref{fig:compare-syn}. For the dynamic parts, i.e.,  the two hands, and their
static parts, i.e., the head and body, our results are both clearer and more accurate. V4D still exhibits some blurriness, while HexPlane and TiNeuVox fall far behind in terms of both clarity and overall quality compared to our results. Our method can better capture the high-frequency information of the underlying dynamic scene thus producing the finer and clearer details.  
}

\begin{figure*}[t]
    \centering
    \includegraphics[width=\linewidth]{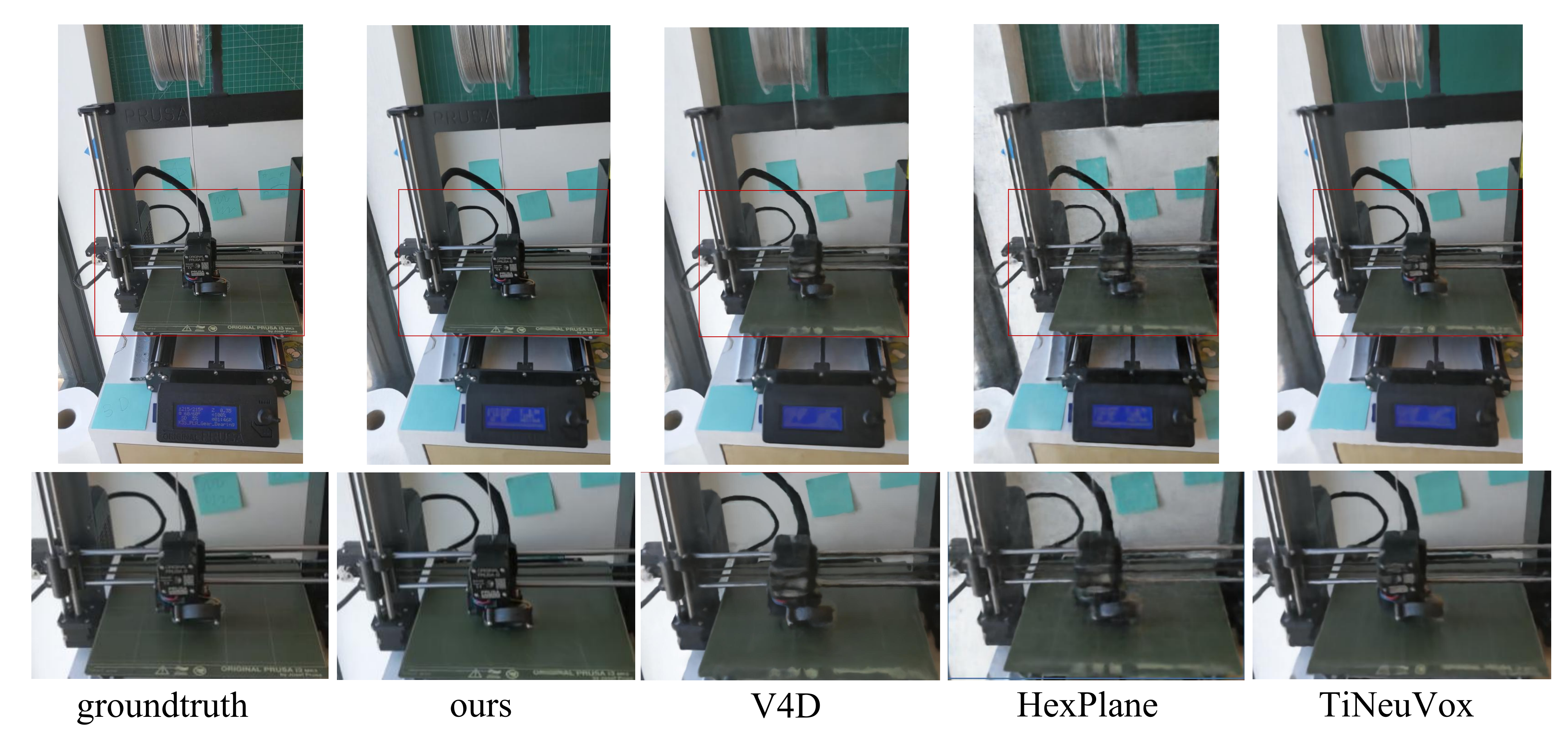}
    \caption{Vsiual comparsion on the real-world dataset HyerNeRF~\cite{park2021hypernerf}. Ours can see the clear text while others fail. Zoom in to see more details.} 
    \label{fig:compare-real}
\end{figure*}

\begin{table}
  \centering
  \caption{Quantantiive comparison on the real-world datset HyperNeRF~\cite{park2021hypernerf}. The best results are in boldface.
  }
  \label{tab:hyper-nerf works compare}
  \begin{tabular}{@{}lcccccccccccccccc@{}}
    \toprule
    Method & PSNR$\uparrow$ & MS-SSIM$\uparrow$ \\
    \midrule
    Nerf\cite{mildenhall2021nerf} &20.1&0.745 \\
    NV\cite{lombardi2019neural} &16.9&0.571 \\
    Nerfies\cite{park2021nerfies} &22.2&0.803 \\
    HyperNerf\cite{park2021hypernerf} &22.4&0.814 \\
    TiNeuVix-S\cite{fang2022fast} &23.4&0.813 \\
    TiNeuVox-B\cite{fang2022fast} &24.3&\textbf{0.837} \\
    HexPlane\cite{cao2023hexplane} &24.0& 0.807\\
    V4D\cite{gan2023v4d} &24.8&0.832 \\
    SLS4D(ours) &\textbf{25.2}& 0.832\\
    \bottomrule
  \end{tabular}
\end{table}

\tj{Following related papers, we report both PSNR and MS-SSIM~\cite{zhang2018unreasonable} on the HyperNeRF~\cite{park2021hypernerf} dataset.}
From the table \ref{tab:hyper-nerf works compare}, it is evident that we have achieved state-of-the-art results in terms of PSNR and the second-best results for MS-SSIM compared to previous works.
While we have discussed the TiNeuVox\cite{fang2022fast} and V4D\cite{gan2023v4d} above, the Nerfies\cite{park2021nerfies} and HyperNeRF\cite{park2021hypernerf} use time-flow-based latent codes, which may need more generality compared to our method. Additionally, using naive MLP instead of the feature space can result in slightly inferior rendering quality~\cite{fang2022fast,cao2023hexplane,gan2023v4d}. 
\fqy{Some visual comparison results are illustrated in Fig.~\ref{fig:compare-real}.
Compared to previous works, our method allows us to see the text clearly in both the dynamic and the static parts, while the details of other works are vague.}

\tj{\tj{More visual results on both the synthetic and real-world datasets can be found in the supplementary material.}}

\begin{figure*}[t]
    \centering
    \includegraphics[width=\linewidth]{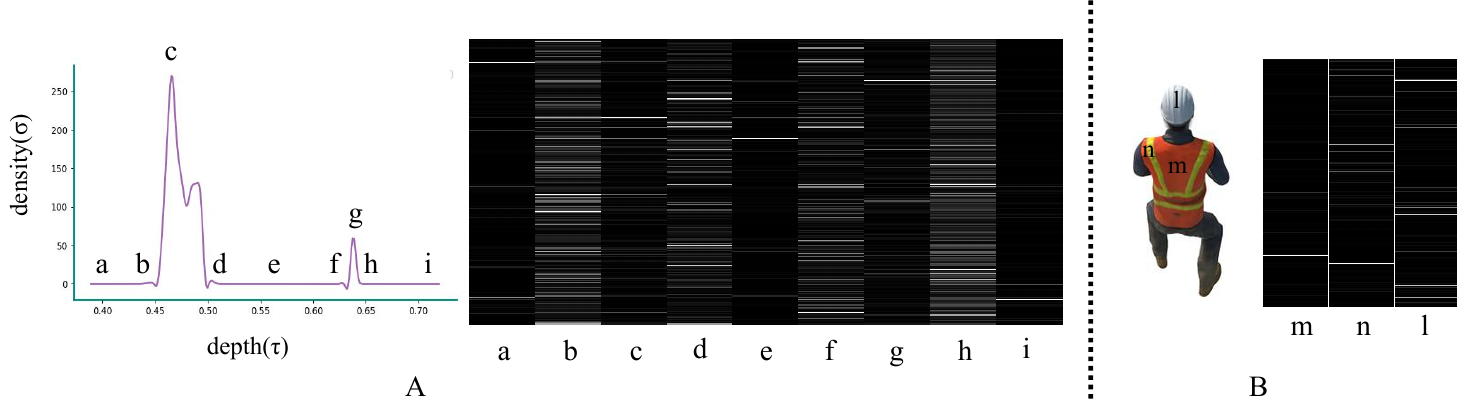}
    \caption{
    Visualization of weights on latent codes.
    A: We sample 9 different points along a ray and plot their weights on different latent codes (along the vertical axis).
    B: We select 3 points on the surface of the dynamic 3D object,  and plot their weights on different latent codes (along the vertical axis).
    A brighter value indicates higher weights.}
    \label{fig:weight feature}
\end{figure*}

\begin{table*}[t]
  \centering
  \caption{Ablation results on the dynamic synthetic dataset of D-NeRF~\cite{pumarola2021d}: $nd$ indicates not using the deformation fields. $nf$ indicates not using the time slots in the deformation fields. $nr$ indicates not using the total variance loss(TV loss). $co$ indicates using the cyclic cosine decay. $eo$ indicates using exponential decay. $ge$ indicates only using gelu as the interpolator decoder. $ml$ indicates only using MLP as the interpolator decoder.}
  \label{tab:d-nerf ablations}
  \begin{tabular}{@{}lcccccccccccccccc@{}}
    \toprule
    Method & deformation fields & time slots & TV loss  & lr decay function & interpolator decoder &PSNR$\uparrow$ & SSIM$\uparrow$ & LPIPS$\downarrow$ \\
    \midrule
    SLS4D$_{nd}$ &            &            &            & cos*exp&GeGLU& 32.56 & 0.971 & 0.038\\
    SLS4D$_{nf}$ & \checkmark &            &            & cos*exp&GeGLU& 34.26 & 0.979 & 0.027\\ 
    SLS4D$_{nr}$ & \checkmark &\checkmark  &            & cos*exp & GeGLU & 33.40 & 0.977 & 0.034\\ 
    SLS4D$_{co}$ & \checkmark &\checkmark  & \checkmark & cos    &GeGLU& 33.17 & 0.975 & 0.043\\ 
    SLS4D$_{eo}$ & \checkmark &\checkmark  & \checkmark & exp    &GeGLU& 34.48 & 0.980 & 0.026\\ 
    SLS4D$_{ge}$ & \checkmark &\checkmark  & \checkmark & cos*exp&GeLU& 34.44 & 0.979 & 0.027\\
    SLS4D$_{ml}$ & \checkmark &\checkmark  & \checkmark & cos*exp&MLP& 34.55 & 0.980 & 0.026\\
    SLS4D        & \checkmark &\checkmark  & \checkmark & cos*exp&GeGLU& 34.84 & 0.981 & 0.025 \\
    \bottomrule
  \end{tabular}
\end{table*}

\subsection{Discussion on Latent Space Representation}
\label{sec:disccus}
Compared to traditional bilinear or trilinear interpolation in the dense and local grid, our implicit interpolation in the latent space introduces new properties.
\tj{To demonstrate the sparsity of the learned radiance fields and} \fqy{the adaptive allocation of weights in our latent space feature interpolation,} \tj{we conducted the following two experiments.}

\paragraph{Test A} For a ray passing through the reconstructed scene, we illustrate the density along the ray in Fig.~\ref{fig:weight feature}(A, left), which rises and falls, reaching its peak at the surface of the object.
\tj{We also illustrate the latent space representation (i.e., weight to each latent code) of each sampled point (a-i).
We find that only the interval $[0.45,0.50]\cup[0.63,0.65]$ has a non-zero density, which indicates the sparsity of the 3D space, where only a very small portion is non-empty.
As we can see clearly from the latent representation, the points can be roughly classified into three categories, i.e., (a, e, i), (c, g) and (b, d, f, h), which are exactly the points of low density, high density and others, respectively.
This implies that our latent space can effectively and adaptively represent the 4D scene by leveraging the spatial sparsity nature of deformation fields and radiance fields.
}

\paragraph{Test B} As shown in Fig.~\ref{fig:weight feature}(B), for regions with complex texture and geometry, such as the hat (l) and shoulder (n), the position is affiliated with more latent codes indicated by the weights. In contrast, for regions with simpler textures, like the back (m), there are fewer latent codes attended to the location. This indicates that our method can adapt to the complexity of local textures and geometry, automatically utilizing different numbers of latent codes to represent different regions.

\subsection{Ablations and Analysis}
To justify the choices of modules in our method, we conduct ablation experiments on the D-NeRF\cite{pumarola2021d} dataset.
The results are shown in Table~\ref{tab:d-nerf ablations}.

\paragraph{Deformation Fields} We first attempt to test our method not using the deformation network. The absence of decoupling between time and space in dynamic 3D scenes poses challenges in extracting information for reconstruction, resulting in a decline in the quality of our reconstruction results, denoted as SLS4D$_{nd}$. This experiment illustrates the robust learning capability of our latent space representation in dynamic 3D space. Even without time decoupling, it outperforms all dynamic 3D reconstruction methods except V4D~\cite{gan2023v4d}.

Afterward, we attempt to incorporate a classical deformation fields, denoted as SLS4D$_{nf}$, similar to D-NeRF, a 7-layer MLP with the input of a 4D coordinate$(x,y,z,t)$ and the output of a 3D coordinate transformation$(\Delta x, \Delta y, \Delta z)$. Time slots play a crucial role in ensuring the smoothness of dense temporal information of deformation. In other words, it ensures a smooth transition while attempting to separate the deformation fields for each timestamp. Removing the time slots results in a degradation of the rendering quality.

We also test a pure time slot without TV loss, denoted by SLS4D$_{nr}$. This approach lacks supervision for temporal smoothness, resulting in poorer performance.

\paragraph{Learning Rate Decay Function}

We endeavor to optimize our SLS4D model using the cosine learning rate decay function, denoted as SLS4D$_{co}$. While this approach enables the escape from local optima, experimental results suggest it is more unstable and susceptible to sub-optimal learning outcomes.

We also conducted experiments not using the exponential decay learning rate function, denoted as SLS4D$_{eo}$. Although this approach ensures stable and continuous improvement, it may encounter challenges in getting stuck in local optima, resulting in sub-optimal outcomes.

\paragraph{Interpolator Decoder}

We also performed experiments using the basic GeLU instead of GeGLU in the interpolator decoder, denoted as SLS4D$_{ge}$. The GeGLU, instead of GeLU, is a bilinear GLU that can offer higher fitting capacity, leading to better results.

Directly using MLP for the interpolator decoder is also attempted, denoted as SLS4D$_{ml}$. The shape of GeLU is similar to the cumulative distribution function (CDF) of the normal distribution, making it more suitable for modeling complex nonlinear relationships than MLP.

\section{Conclusion and Limitations}

We have presented a novel approach for 4D novel-view synthesis, called SLS4D, which represents the 4D space in a sparse, compact and informative latent feature space, reducing the overall network parameters, and achieving higher quality of 4D novel-view synthesis. The network parameters are approximately $6\%$ of the latest work, V4D~\cite{gan2023v4d}.  SLS4D is the first method to use latent space representation for dynamic neural radiance fields and demonstrates the significant potential of implicit interpolation in neural radiance fields. 

% Although our training time is  the running speed may be relatively slow. Therefore, accelerating this implicit interpolation strategy is a research question worth investigating. Additionally, for inputs with excessively long-duration, a small number of parameters may struggle to fit the data adequately. Therefore, how to address fitting for extremely long-duration motion scenes is also a research question worth exploring.
Despite our method exhibiting shorter training times compared to V4D~\cite{gan2023v4d}, considerable optimization potential still exists. A strategy incorporating coordinate-mapped feature masks to minimize the computation burden can be used to improve computation speed. Additionally, for inputs with excessively long-duration, a small number of parameters may struggle to fit the data adequately. A growing NeRF\cite{yang2022recursive} model can be used to solve this problem by adaptively splitting the spatial dimensions along the time dimension.

% \qcx{Please add Limitation and Feature work in this paragraph.}
{
    %\small
    \bibliographystyle{IEEEtran}
    \bibliography{main}
}


\newpage

\begin{IEEEbiography}[{\includegraphics[width=1in,height=1.25in,clip,keepaspectratio]{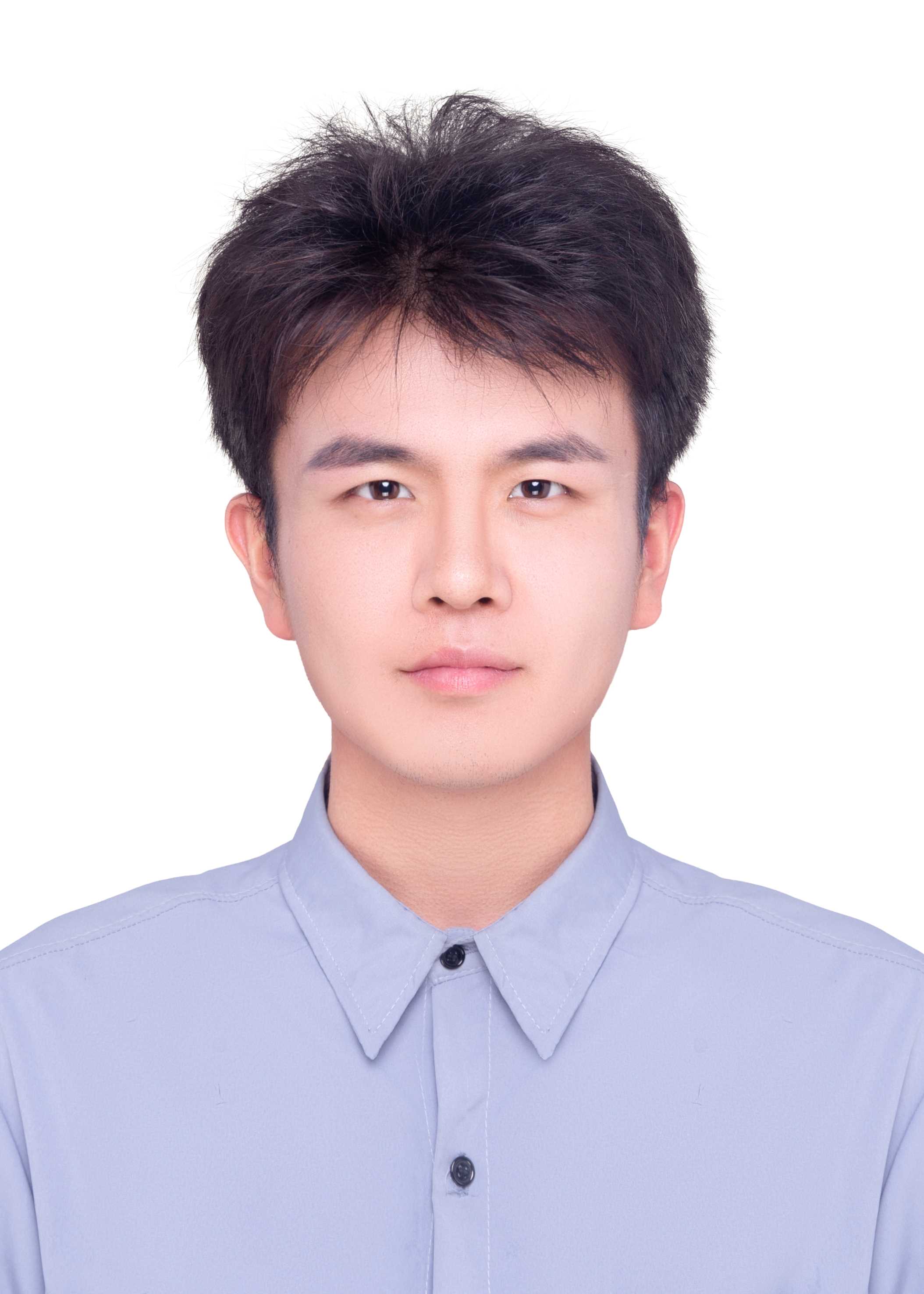}}]
    {Qi-Yuan Feng} received his bachelor's degree in computer science from Tsinghua University in 2022. He is currently a PhD candidate in the Department of Computer Science and Technology, Tsinghua University. His research interests include 3D reconstruction and geometry processing.

\end{IEEEbiography}

\vspace{11pt}

\begin{IEEEbiography}[{\includegraphics[width=1in,height=1.25in,clip,keepaspectratio]{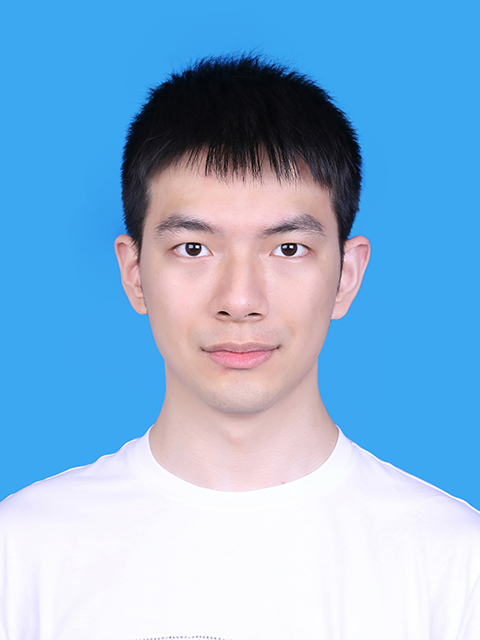}}]{Hao-Xiang Chen} received his bachelor’s degree in computer science from JiLin University in 2020. He is currently a PhD candidate in the Department of Computer Science and Technology, Tsinghua University. His research interests include 3D reconstruction and 3D computer vision.
\end{IEEEbiography}

\vspace{11pt}

\begin{IEEEbiography}[{\includegraphics[width=1in,height=1.25in,clip,keepaspectratio]{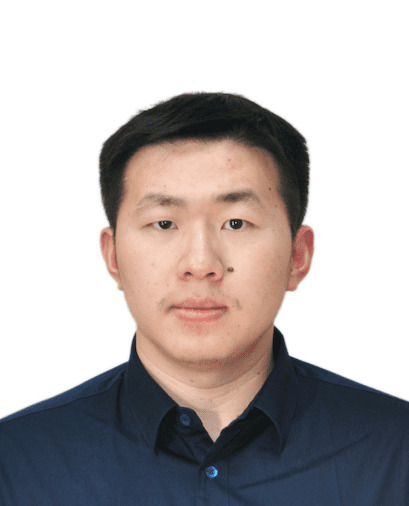}}]{Qun-Ce Xu} is a postdoctoral researcher in the Department of Computer Science and Technology at Tsinghua University, Beijing, China. He received his Ph.D. degree from the University of Bath, UK, in 2021. His research interests include geometric learning, geometry processing and shape generation.
\end{IEEEbiography}

\vspace{11pt}

\begin{IEEEbiography}[{\includegraphics[width=1in,height=1.25in,clip,keepaspectratio]{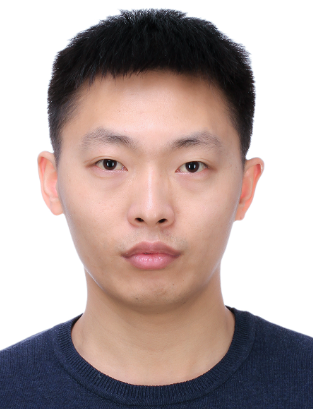}}]{Tai-Jiang Mu} is currently an assistant researcher in the Department of Computer Science and Technology, Tsinghua University, where he received his bachelor’s and doctor’s degrees in 2011 and 2016, respectively. His research interests include computer graphics, visual media learning, 3D reconstruction and 3D understanding.
\end{IEEEbiography}

\vfill

\end{document}